\documentclass[10pt,twocolumn,letterpaper]{article}

\usepackage{iccv}
\usepackage{times}
\usepackage{epsfig}
\usepackage{graphicx}
\usepackage{amsmath}
\usepackage{amssymb}
\usepackage{url}
\usepackage{multirow}
\usepackage{capt-of,etoolbox}
\usepackage{caption}
\usepackage{lipsum }
\usepackage[ruled,vlined]{algorithm2e}
\usepackage{bm}
\usepackage{authblk}

\usepackage[pagebackref=true,breaklinks=true,letterpaper=true,colorlinks,bookmarks=false]{hyperref}

\iccvfinalcopy 

\def\iccvPaperID{445} 

\ificcvfinal\fi
\begin{document}

\title{Automatic Layer Separation using Light Field Imaging}

\author[1]{Qiaosong Wang}
\author[1]{Haiting Lin}
\author[2]{Yi Ma}
\author[3]{Sing Bing Kang}
\author[1]{Jingyi Yu}
\affil[1]{University of Delaware, Newark, DE, USA}
\affil[2]{ShanghaiTech University, Shanghai, China}
\affil[3]{Microsoft Research, Redmond, WA, USA}
\maketitle

\begin{abstract}
We propose a novel approach that jointly removes reflection or translucent layer from a scene and estimates scene depth. The input data are captured via light field imaging. The problem is couched as minimizing the rank of the transmitted scene layer via Robust Principle Component Analysis (RPCA). We also impose regularization based on piecewise smoothness, gradient sparsity, and layer independence to simultaneously recover 3D geometry of the transmitted layer. Experimental results on synthetic and real data show that our technique is robust and reliable, and can handle a broad range of challenging layer separation problems.
\end{abstract}

\section{Introduction}

Reflections and transparency are prevalent in real scenes, and are typically viewed as undesirable. Unfortunately, it is non-trivial to remove them. The observed image $I$ can be generally modeled as a linear combination of a transmitted layer $T$ (which contains the scene of interest) and a secondary layer $S$ (which contains the reflection or transparency). Typical examples include a picture behind a glass cover and a scene blocked by a sheer curtain. Extracting $S$ from $I$ is a problem that is inherently ill-posed: we have two unknowns $T$ and $S$ but only one equation. To make this underconstrained problem more tractable, existing solutions either impose additional priors (e.g., through user inputs or spatial regularization) ~\cite{levin2007user,levin2004separating} or use more constraints (e.g., by capturing more photographs) ~\cite{szeliski2000layer,tsin2006stereo,li2013exploiting,guo2014robust}.

   \begin{figure}
   \begin{center}
   \includegraphics[height=5.3 cm]{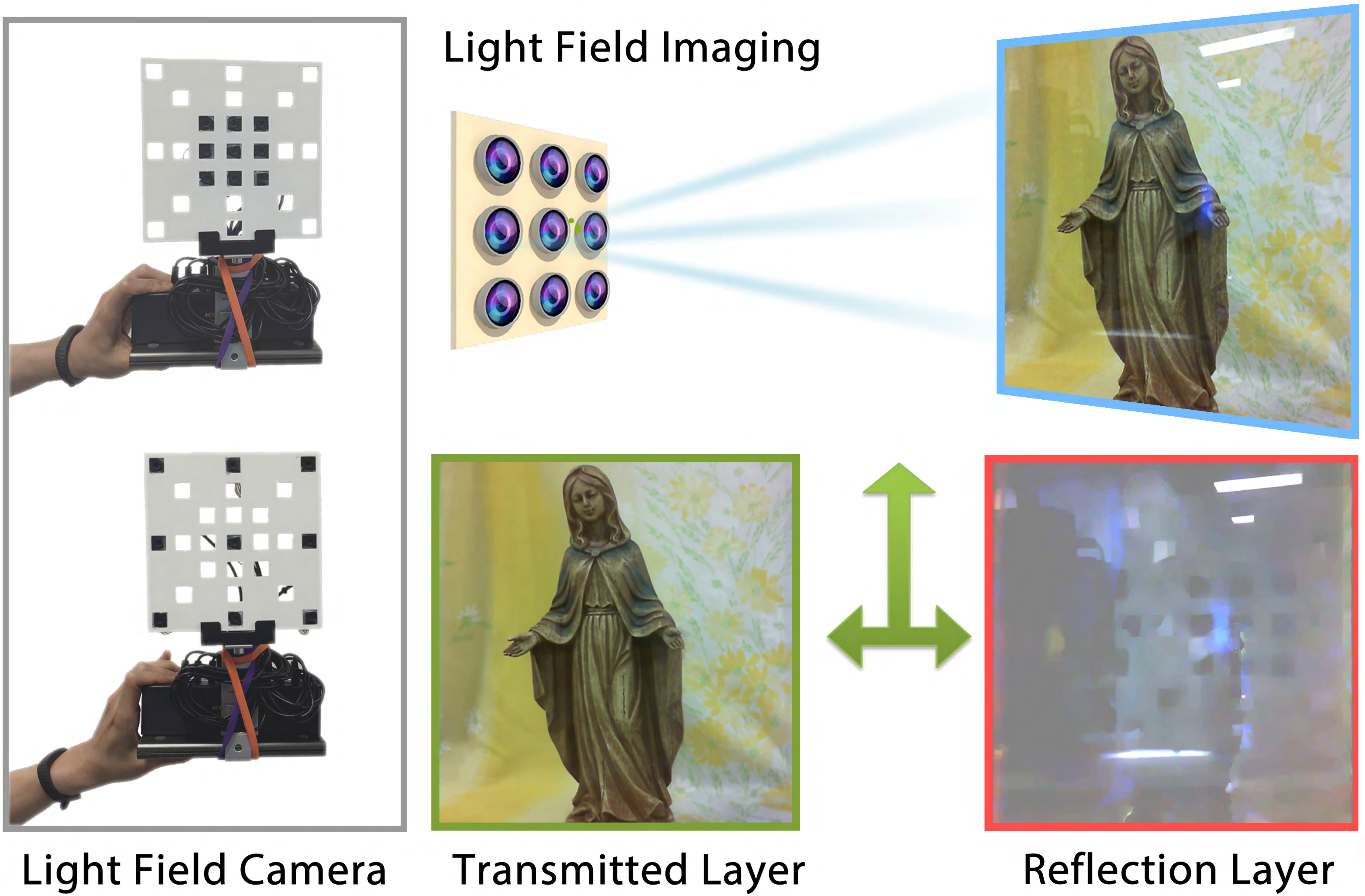}
   \end{center}
   \caption{\emph{Left:} Our portable camera array with a reconfigurable baseline. \emph{Right:} We demonstrate how to exploit such a light field camera for layer separation tasks.}
   { \label{fig:sample_lf}}
   \end{figure}

In this paper, we present a new computational imaging solution by exploiting emerging light field imaging techniques. A light field (LF) captures an array of images from a grid of viewpoints. It can be viewed as a single-shot multi-view imaging system. The multi-view attribute enables reliable depth estimation ~\cite{heber2014shape,wanner2013variational,kim2013scene,ChenCVPR2014} that eliminates the need of homography assumption in ~\cite{szeliski2000layer,gai2012blind,tsin2006stereo,guo2014robust}. Our technique begins with estimating an initial disparity map using SIFT flow ~\cite{liu2011sift}. We then warp all LF views to the reference view (in our case, the central camera) to form an image stack. We show that the image stack exhibits low-rank property, and we apply Robust Principle Component Analysis (RPCA) for simultaneous layer separation and disparity refinement.

A unique advantage of our LF-based solution is that we can represent scene geometry as a single disparity map under which the resulting warped image stack will be low-rank. In contrast, the warped image stack in previous multi-view approaches is only low-rank when scene geometry is planar (via homographic warping on the cropped common region) and they can break down on complex scenes (Fig.~\ref{fig:synthetic} and \ref{fig:depth}). We conduct experiments on both synthetic and real data. In particular, we construct a $3\times3$ mini LF array that is portable and can be controlled by a single tablet. Results on static and dynamic scenes show that our technique is robust and reliable and can handle a broad range of challenging layer separation problems.


\section{Related Work}
The problem of image layer separation is ill-posed, and typically relies on additional priors or constraints. Earlier approaches rely on user inputs to provide priors on the two layers. Levin et al.~\cite{levin2007user} develop a user-assisted system to label image gradients to one of the two layers. An automatic method can then be used to search for a decomposition that minimize the total amount of edges and corners, using a database of natural image patches~\cite{levin2004separating}.

To automate the layer separation process, more recent techniques use multiple images, either from a fixed viewpoint with varying camera settings (such as flash, focus, and polarization), or from multiple viewpoints through the use of a hand-held camera~\cite{hirschmuller2008stereo,szeliski2000layer,gai2012blind,tsin2006stereo,sinha2012image,li2013exploiting,guo2014robust}. In the case of the fixed viewpoint approach, ~\cite{farid1999separating,schechner2000polarization,kong2011high} exploit the effect of reflection under different rotation angles of a polarizer. Agrawal \emph{et al.}~\cite{agrawal2005removing} show how a flash/no-flash image pair can be used to remove both reflections and highlights through gradient filtering and integration. Schechner \emph{et al.}~\cite{schechner2000separation} propose to vary the focus of the camera for eliminating reflection artifacts. The use of different modes of capture is complementary to our technique.

Methods for separating layers using multiple-viewpoint images are based on the intuition that the transmitted layer and reflection undergo different motions under changing views. Szeliski \emph{et al.}~\cite{szeliski2000layer} propose to separate the two layers by estimating global and local motions. Gai \emph{et al.} ~\cite{gai2012blind} study the statistics of natural images to extract both the motion of the two layer motions and their mixing coefficients. In a similar vein, Tsin \emph{et al.}~\cite{tsin2006stereo} assume locally planar motion and require dense image capture to estimate both the depth and appearance of each layer through EPI analysis. Sinha \emph{et al.}~\cite{sinha2012image} speed up the process by adopting piecewise planar scene models and extends the semi-global matching ~\cite{hirschmuller2008stereo} for reliable layer separation. More recently, Guo \emph{et al.}~\cite{guo2014robust} correlate all images through homography and then conduct low-rank decomposition to effectively separate the reflection layer from the transmitted layer. Although these techniques are effective, the requirement of capturing multiple and often many images of scene from different viewpoints, and hence time instances, significantly limits their applicability. Further, there is an implicit assumption that the scene is mostly planar and can be rectified via a homography.

We seek a single-shot solution through LF imaging. The concept of LF imaging can be traced by integral photography by Lippmann~\cite{Lippermann1908} in which a lenslet array is used to emulate acquisition of multiple viewpoints~\cite{adelson1992single,ng2005light,lumsdaine2009focused}. Hand-held plenoptic cameras are now commercially available~\cite{lytro} and mobile camera arrays \cite{Pelicanimaging,Lightstartup} will be on the market soon. In our experiments, we use a mini LF camera array to support on-site acquisition. Techniques that capitalize on the availability of such cameras include~\cite{HeberICEMM2013,heber2014shape} (variational shape from LF data), \cite{YuICCV2013} (line assisted stereo matching), \cite{taodepth2014} (depth estimation of glossy surfaces), and \cite{ChenCVPR2014} (robust stereo matching).

  \begin{figure}
   \begin{center}
   \includegraphics[height=6.50 cm]{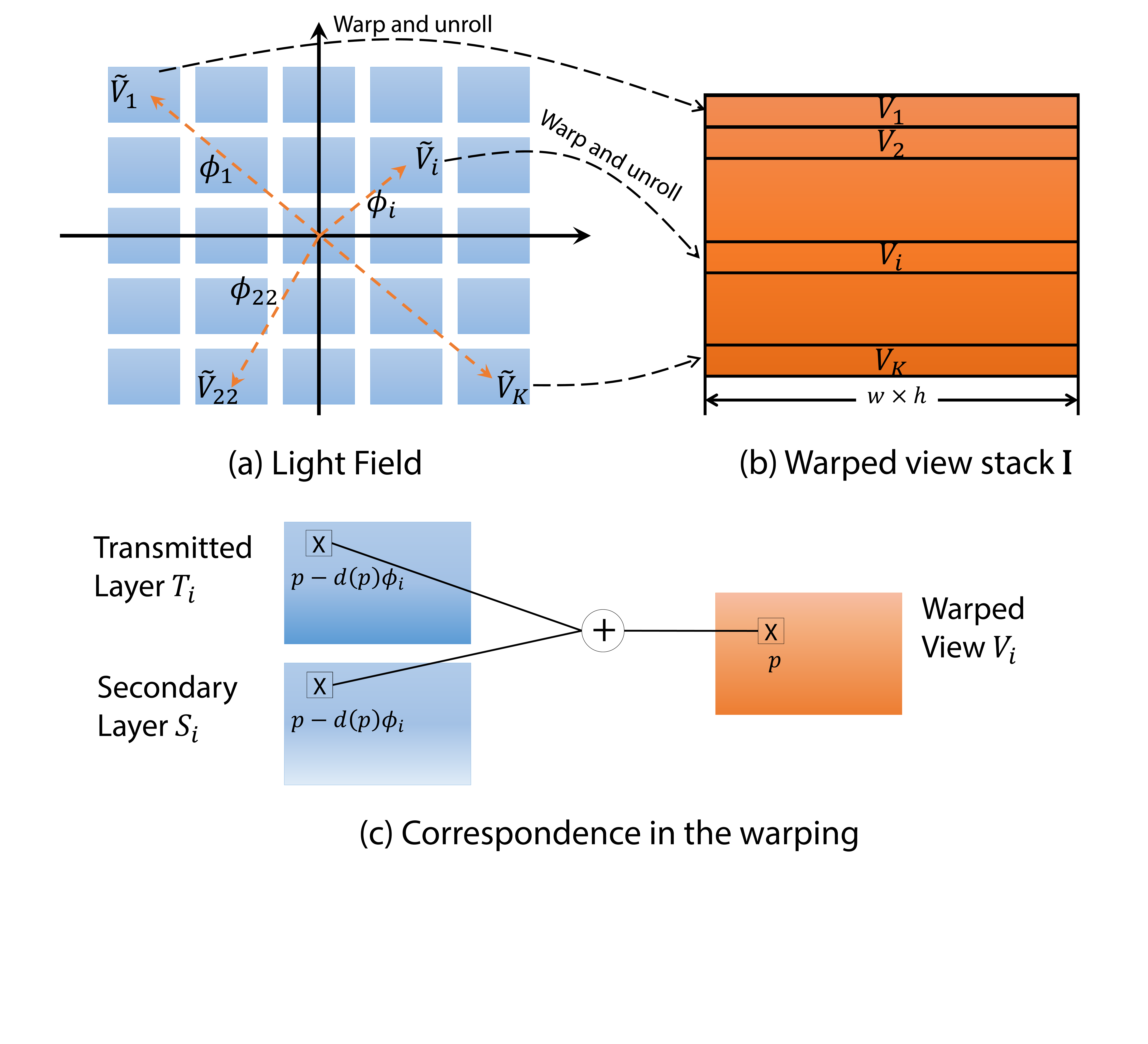}
   \end{center}
   \caption{Warping light field views to an image stack. Every light field view (a) is unrolled as a row vector and stacked into a matrix (b) using the disparity map. We decompose it into the transmitted and secondary matrices (c).
   }{ \label{fig:lf_warping}}
   \end{figure}


\section{Problem Formulation}
In our work, we capture the LF of the scene (transmitted layer) that has been superimposed with a secondary layer (e.g., reflection).
The inputs are LF images from different viewpoints, and we take the central view as the reference view. Our goal is to separate the layers for the reference view by exploring redundant information that is available from the other views. To account for scene appearance in all the views, we estimate the disparity map of the transmitted layer; this map is used to align all the LF views with respect to the reference to facilitate layer separation. The disparity map estimation and layer separation steps are done iteratively.

We first explain our notations. Our LF consists of a 2D grid of $K = N\times N$ viewpoints, with each image having a resolution of $w\times h$. The $i$-th 2D sub-aperture image is unrolled as a 1D image vector $\widetilde{V}_{i}$, $i\in\{1,2,...,K\}$; the term $\phi_{i}$ maps index $i$ to its position within the 2D image grid. We assume the images are uniformly sampled horizontally and vertically with an identical baseline and $d$ represents the disparity map of the reference view with respect to its one-hop neighbor view. We use $V_i$ to represent the warped result from $\widetilde{V}_{i}$ to the reference using $d$. As with $\widetilde{V}_{i}$, $V_i$ and $d$ are also unrolled into 1D row vectors $\in \mathbb{R}^{1\times hw}$. Given $d$, we can compute $V_i$'s and stack them to form matrix $I \in \mathbb{R}^{K \times hw}$. The warped LF images will now contain the warped transmitted and secondary layers: $V_i = T_i + S_i$. We can similarly stack all $T_i$ and $S_i$ into two matrices $T$ and $S \in \mathbb{R}^{K\times hw}$.
Fig.~\ref{fig:lf_warping} illustrates the warping process.

Our goal is to recover $T$, $S$, and $d$ from a single equation $I = T + S$. Since this problem is ill-posed, we need to impose additional constraints as in~\cite{guo2014robust}. First, the transmitted layer should be the same after disparity warping to the reference view, and therefore should be of low rank. In contrast, the warped secondary layer should have pixel-wise low coherence across views because they are warped using the disparity of the transmitted layer rather than their own disparity map, and therefore $S$ should be sparse.
In addition, the transmitted and secondary layers should be independent and their gradients sparse. Putting all these together, we formulate the layer separation problem as energy minimization:
\begin{equation}\label{eq:eq1}
\begin{aligned}
& \underset{T,S,d,\omega}{\text{minimize}} \quad
rank\left( T\right)   + \\
 & \lambda_{1}\|DT\odot DS\|_{0}+
 \lambda_{2}\|DI-DT-DS\|_{F}^{2}  \\
 & + \lambda_{3} \|DT\|_{0}+ \lambda_{4} \|DS\|_{0} \\
 & + \lambda_{5}  \|  d-\omega  \|_{1} +\lambda_{6} \|D\omega \|_{1} \\
& \text{subject to} \quad  I =T+S; T\succeq 0; S\succeq 0 ,
\end{aligned}
\end{equation}
where $\| \cdot \|_{0}$, $\| \cdot \|_{1}$, and $\| \cdot \|_{F}$ are $\ell ^{0}$, $\ell ^{1}$, and Frobenius norm respectively, $\omega$ is an intermediate variable for refining the disparity map $d$, $\odot$ represents the element-wise multiplication, and $D$ is the finite difference operator applied to an image on both x and y direction.

In this formulation, the first term forces the rank of matrix $T$ to be low. The second and third terms force the gradients of the two layers to be mutually independent. The fourth and fifth terms imposes the sparse gradient prior on natural images. The last two terms employ $\ell ^{1}$-TV to refine the disparity map $d$. We choose $\ell ^{1}$-TV instead of $\ell ^{2}$-TV as the regularization term for two reasons. First, a disparity map is largely piecewise constant. Second, the $\ell ^{1}$ norm measure $\|  d-\omega  \|_{1}$ is commonly used for evaluating the percentage of bad pixels on disparity maps \cite{scharstein2002taxonomy}. Therefore, $\|  d-\omega  \|_{1}$ can be interpreted as the convexification of bad pixel percentage in $d$. We further impose hard constraints that $T$ and $S$ be non-negative ($T\succeq 0, S\succeq 0$). The optimization problem, however, is NP-hard. We follow \cite{candes2011robust} to solve an alternative convex relaxation problem:
\begin{equation}\label{eq:eq2}
\begin{aligned}
& \underset{T,S,d, \omega}{\text{minimize}} \quad
\left\| T\right\|_{*}  +\\
 & \lambda_{1}\|DT\odot DS\|_{1}+
 \lambda_{2}\|DI-DT-DS\|_{F}^{2}  \\
 & + \lambda_{3} \|DT\|_{1}+ \lambda_{4} \|DS\|_{1} \\
 & + \lambda_{5}  \|  d-\omega  \|_{1} +\lambda_{6} \|D\omega \|_{1} \\
& \text{subject to} \quad  I =T+S; T\succeq 0; S\succeq 0
\end{aligned}
\end{equation}
where nuclear norm $\left\| \cdot\right\|_{*}$ replaces the $rank$ function and $\ell ^{1}$ norm replaces $\ell ^{0}$ norm in Eq.~\ref{eq:eq1}.

The new formulation now allows convex optimization. However, the 3D-warping function $I(d)$ is still highly non-linear. In order to linearize the warping function, we further formulate ${V}_{i}$ as:
\begin{equation}\label{eq:eq3}
\begin{split}
{V}_{i}(p)=\widetilde{V}_{i}(p-d(p)\phi_{i} ) ,
\end{split}
\end{equation}
where $p$ is the image pixel coordinate. In order to convert the objective function into a convex model, we follow \cite{heber2014shape} to linearize the warped images using first order Taylor approximation on disparity $d^{(t)}$ at iteration $t$. For each image, we have:
\begin{equation}\label{eq:eq4}
V_{i}^{\, (t+1)}(p)
\approx
\widetilde{V}_i(p-d^{(t)}(p)\phi_{i})+ (d^{(t+1)}(p)-d^{(t)}(p))\cdot\widehat{J}_{i} ,
\end{equation}
where $\widehat{J}_{i} \in \mathbb{R}^{hw\times1}$ is
\begin{equation}\label{eq:eq5}
\widehat{J}_{i}=
    \left| \left| \phi_{i}\right| \right|  \nabla_{-\frac {\phi _{i}} {\left\| \phi _{i}\right\| }} \widetilde{V}(p-d^{(t)}(p)\phi _{i}) .
\end{equation}
Letting $J_{i}=diag(\widehat{J}_{i})$, we rewrite the constraint in Eq.~\ref{eq:eq2} as:
\begin{equation}\label{eq:eq6}
I+\sum _{i=1}^{K}\epsilon_i(\Delta d{J}_{i}) =T+S; T\succeq 0; S\succeq 0 ,
\end{equation}
where $I=I(d^{(t)})$, $\Delta  d =d^{(t+1)}-d^{(t)}$, and $\{\epsilon_i\}$ is the standard basis for
$\mathbb{R}^{K}$. The constraint can be regarded as linearizing the 3D-warping operation with respect to the disparity map $d$.

   \begin{figure*}
   \begin{center}
   \includegraphics[height=6.7 cm]{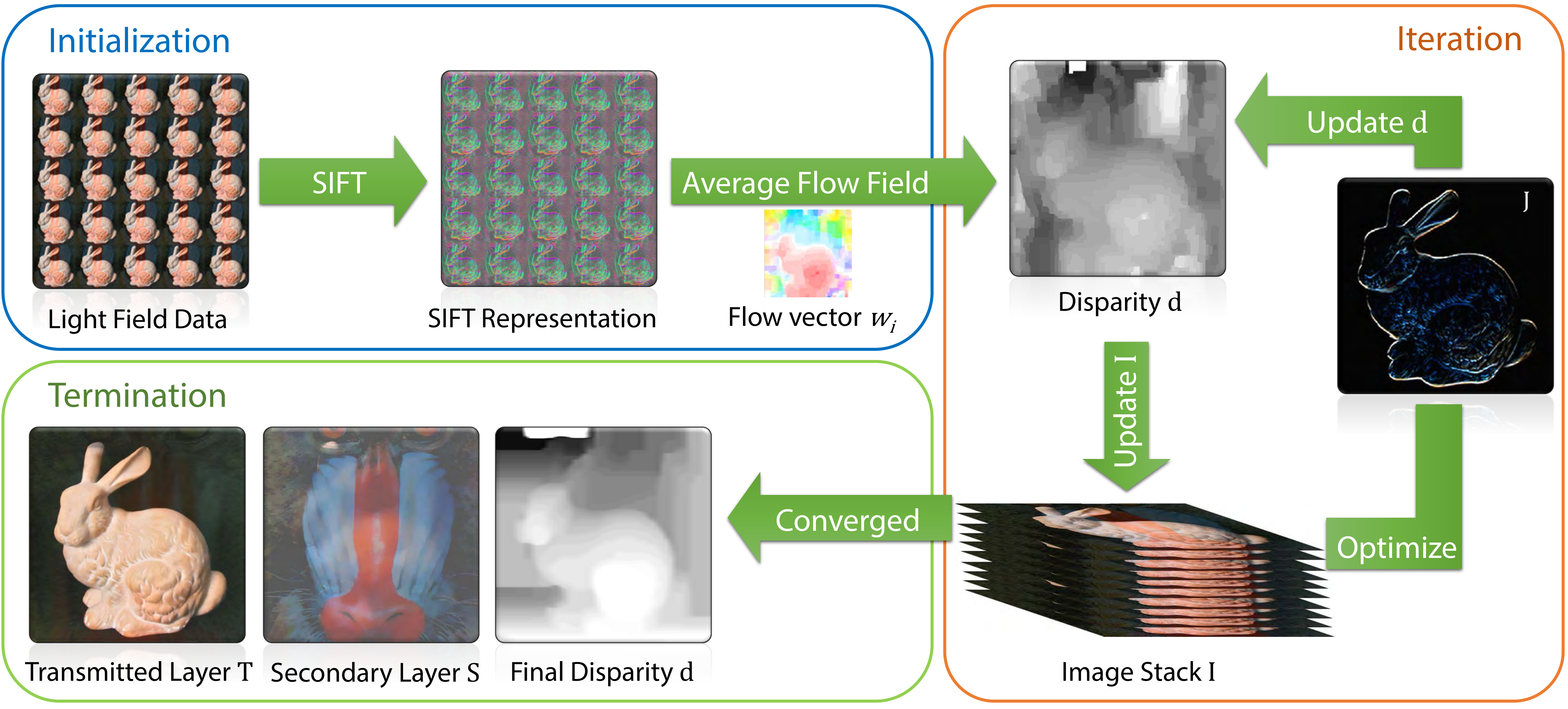}
   \end{center}
    \caption{Illustration of our processing pipeline (see Section \ref{sectionopt} for details).}
   { \label{fig:outline}}
   \end{figure*}


Finally, we combine all priors to simultaneously solve for the transmitted and secondary layers as well as the disparity map by solving the following convex optimization problem:
\begin{equation}\label{eq:eq8}
\begin{aligned}
& \underset{T,S,d,\omega}{\text{minimize}} \quad
\left\| T\right\|_{*}  + \lambda_{1}\|DT\odot DS\|_{1} \\
& + \lambda_{2}\|DI-DT-DS\|_{F}^{2} +
  \lambda_{3} \|DT\|_{1}+ \lambda_{4} \|DS\|_{1} \\
& + \lambda_{5}  \| d-\omega  \|_{1} +\lambda_{6} \|D\omega \|_{1} \\
& s.t. I+\sum _{i=1}^{K}\epsilon_i(\Delta d {J}_{i}) =T+S; T\succeq 0; S\succeq 0 .\\
\end{aligned}
\end{equation}

\section{Optimization}\label{sectionopt}

In this section, we describe how to optimize the objective function defined in Eq. \ref{eq:eq8}. The algorithm is outlined in Algorithm \ref{Algorithm} and illustrated in Fig. \ref{fig:outline}.

\subsection{Initialization}

Our approach starts by warping the sub-aperture images to the center view. Previous studies assume global parametric motion (e.g., homographies~\cite{szeliski2000layer,gai2012blind,guo2014robust}). Despite its computational efficiency and robustness, this approach is unable to handle more complex parallax. In reality, the transmitted layer is unlikely to be planar and a dense 3D reconstruction would be needed for warping the images. Conceptually, we can apply LF stereo matching such as~\cite{wanner2013variational,kim2013scene,ChenCVPR2014}
to first estimate the 3D geometry. However, with the secondary layer corrupting the transmitted layer, direct depth estimation incurs significant errors. 
In our implementation, we use SIFT flow~\cite{liu2011sift} for correspondence, since it has been shown to be effective for registering reflective scenes \cite{li2013exploiting,maeno2013light}.

Similar to the optical flow, SIFT flow only allows descriptors to be matched along the flow vector $w_i(u)=(w_{ix}(u), w_{iy}(u))$ which is composed of the horizontal and the vertical components. 
This fits well to our model since the relative motion between the sub-aperture images and the reference image should approximately follow the flow. The initial disparity $d^{\,0}$ is then obtained by averaging local flows, i.e.,
\begin{equation}\label{eq:flow}
d^{\,0}\left(  u\right) =\dfrac{1}{K} {\sum _{i=1}^{K}\dfrac {w_i(u)^{T}w_i(u)} {w_i(u)^{T}\phi _{i}}} .
\end{equation}

\subsection{Iterative Optimization}

Given the initial disparity estimation, we use the recently proposed Augmented Lagrange Multiplier (ALM) with Alternating Direction Minimizing (ADM) strategy \cite{guo2014robust} to optimize our objective function~\ref{eq:eq8}. Specifically, we can separate the objective into individual sub-problems by introducing five auxilliary variables: $A=T, B=DT, C=DS,E=d-\omega,F=D\omega$.
We also use an intermediate variable $G$ to represent $I+\sum _{i=1}^{K}\epsilon_i(\Delta d{J}_{i})$. Under our formulation, the augmented Lagrangian function can now be represented as:
\begin{equation}\label{eq:ALF}
\begin{aligned}
& \mathcal{L} (T,S,d,\omega,A,B,C,E,F) \\
& = \left\| T\right\|_{*}   + \lambda_{1}\|B \odot C\|_{1}  + \lambda_{2}\|DI-B-C\|_{F}^{2} \\
& + \lambda_{3} \|B\|_{1}+ \lambda_{4} \|C\|_{1}  + \lambda_{5}  \| E  \|_{1} +\lambda_{6} \|F\|_{1} \\
& + \Phi (L_{1}, G-T-S) \\
& + \Phi (L_{2}, A-T)
 + \Phi (L_{3}, B-DT)+ \Phi (L_{4}, C-DS) \\
& + \Phi (L_{5}, E-d+\omega)+ \Phi (L_{6}, F-D\omega) ,\\
\end{aligned}
\end{equation}
where $\Phi (X, Y)=\langle X,Y\rangle+\dfrac {\mu } {2}\left| \left| Y\right| \right| _{F}^{2}$, $\mu$ is a positive scalar, and $L_{1},...,L_{6}$ are Lagrange multipliers. The goal of ALM is to find a saddle point of $ \mathcal{L} (T,S,d,A,B,C,E,F) $, which approximates the solution of the original problem. We adopt the alternating direction method to iteratively solve the subproblems. The solutions and steps for each sub-problems are listed in the Appendix (attached as supplementary material).

Once we obtain the solutions at each iteration, we further update the multipliers as:
\begin{equation}\label{eq:multiplier}
\begin{aligned}
&L_{1}^{t+1}=L_{1}^{t}+\mu^{t}(G^{t+1}-T^{t+1}-S^{t+1})\\
&L_{2}^{t+1}=L_{2}^{t}+\mu^{t}(A^{t+1}-T^{t+1})\\
&L_{3}^{t+1}=L_{3}^{t}+\mu^{t}(B^{t+1}-DT^{t+1})\\
&L_{4}^{t+1}=L_{4}^{t}+\mu^{t}(C^{t+1}-DS^{t+1})\\
&L_{5}^{t+1}=L_{5}^{t}+\mu^{t}(E^{t+1}-d^{t+1}+\omega^{t+1})\\
&L_{6}^{t+1}=L_{6}^{t}+\mu^{t}(F^{t+1}-d\omega^{t+1}) .\\
\end{aligned}
\end{equation}

Algorithm~\ref{Algorithm} shows the complete process. The termination condition is when the change of the objective function between two consecutive iterations is ultra small (0.1 in our experiments). The inner loop terminates when $\|G^{t+1}-T^{t+1}-S^{t+1}\|_{F}\leq\|10^{-4}\|_{F}$ or the maximum number of iterations is reached.

\begin{algorithm}
\caption{Layer separation and Depth Estimation}\label{Algorithm}
\textbf{Input:} Raw Light Field Data $R$ \\
\textbf{Initialize:} $\lambda_{1},...,\lambda_{6}>0. T^{\ 0}=S^{\ 0}=\omega^{\ 0}=A^{\ 0}=B^{\ 0}=C^{\ 0}=E^{\ 0}=F^{\ 0}=0$, $d^{\ 0}=0$, $t=0$, $\mu^{0}>0$, $n>1$\\
\While{$i\leq K$}{
Compute SIFT flow of view $\widetilde{V}_i$ w.r.t. center view $\widetilde{V}_{ref}$; \\
Initialize disparity map $d$;\\
Update $I(d)$ by warping $\widetilde{V}_i$ to center view $\widetilde{V}_{ref}$;\\
}
\textbf{Iteration:}\\
\While{not converged }{
    \While{not converged }{
    Update $A^{\ t+1},B^{\ t+1},C^{\ t+1},E^{\ t+1},F^{\ t+1},\omega^{\ t+1},S^{\ t+1}$,\\
    $T^{\ t+1},\Delta d^{\ t+1}$;\\
    Update $L^{\ t+1}_{1},L^{\ t+1}_{2},L^{\ t+1}_{3},L^{\ t+1}_{4},L^{\ t+1}_{5},L^{\ t+1}_{6}$  via Eq.~\ref{eq:multiplier};\\
    $\mu^{t+1}=n\mu^{t}$;\\
    $t=t+1$;
    }
    Update $d^{\ t+1} = d^{\ t} + \Delta d^{\ t+1}$;\\
    Update $I$;
}
\textbf{Output:} Separated transmitted layer $T$, secondary layer $S$ and disparity map $d$. \\
\end{algorithm}

   \begin{figure*}
   \begin{center}
   \includegraphics[height=7.8cm]{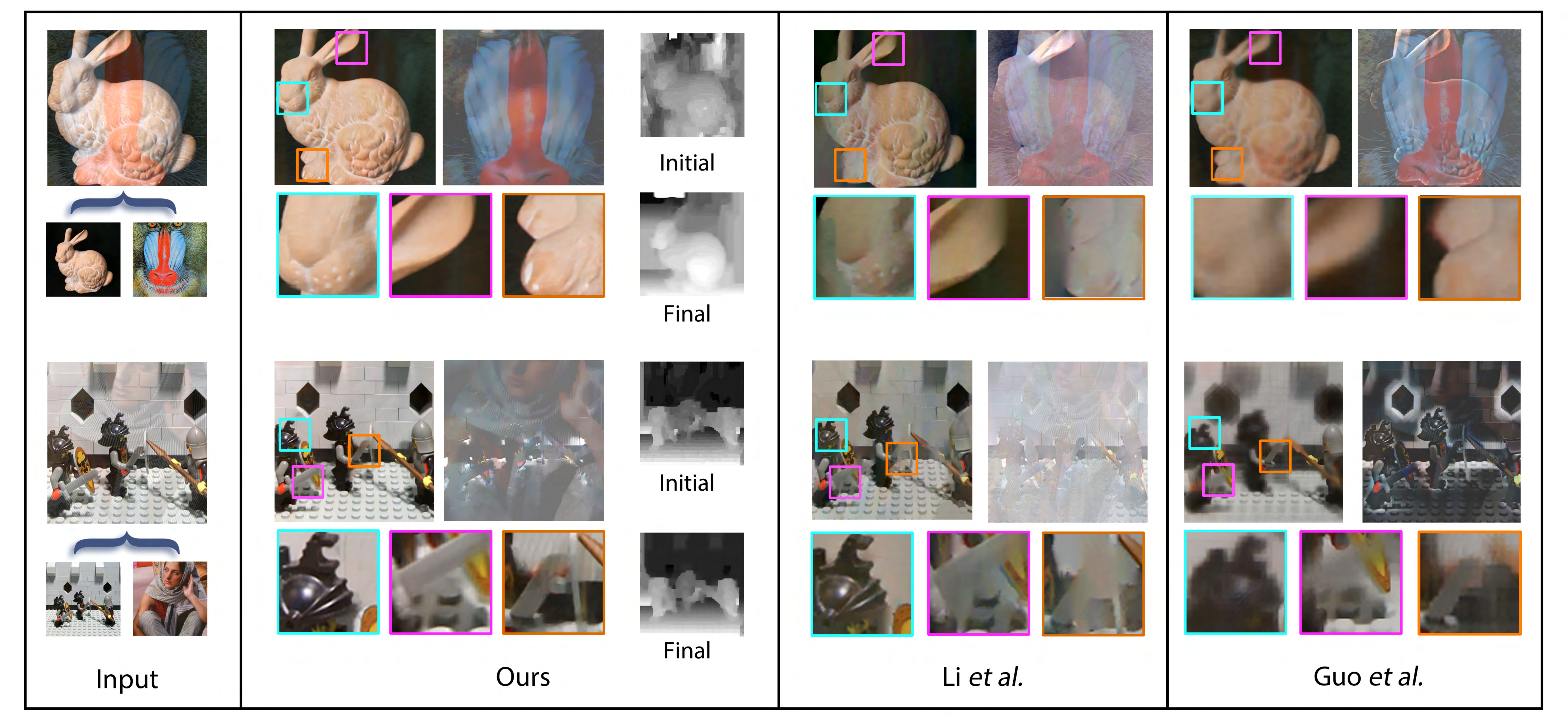}
   \end{center}
   \caption{Results on synthetic data. The recovered secondary layer has been enhanced. Column 1 shows the sample input images; Columns 2 - 4 show results using our technique, \protect\cite{li2013exploiting} and \protect\cite{guo2014robust}. For each technique, we show the recovered transmitted layer and secondary layer with close-up views. }
   { \label{fig:synthetic}}
   \end{figure*}

   \begin{figure}
   \begin{center}
   \includegraphics[height=5.1 cm]{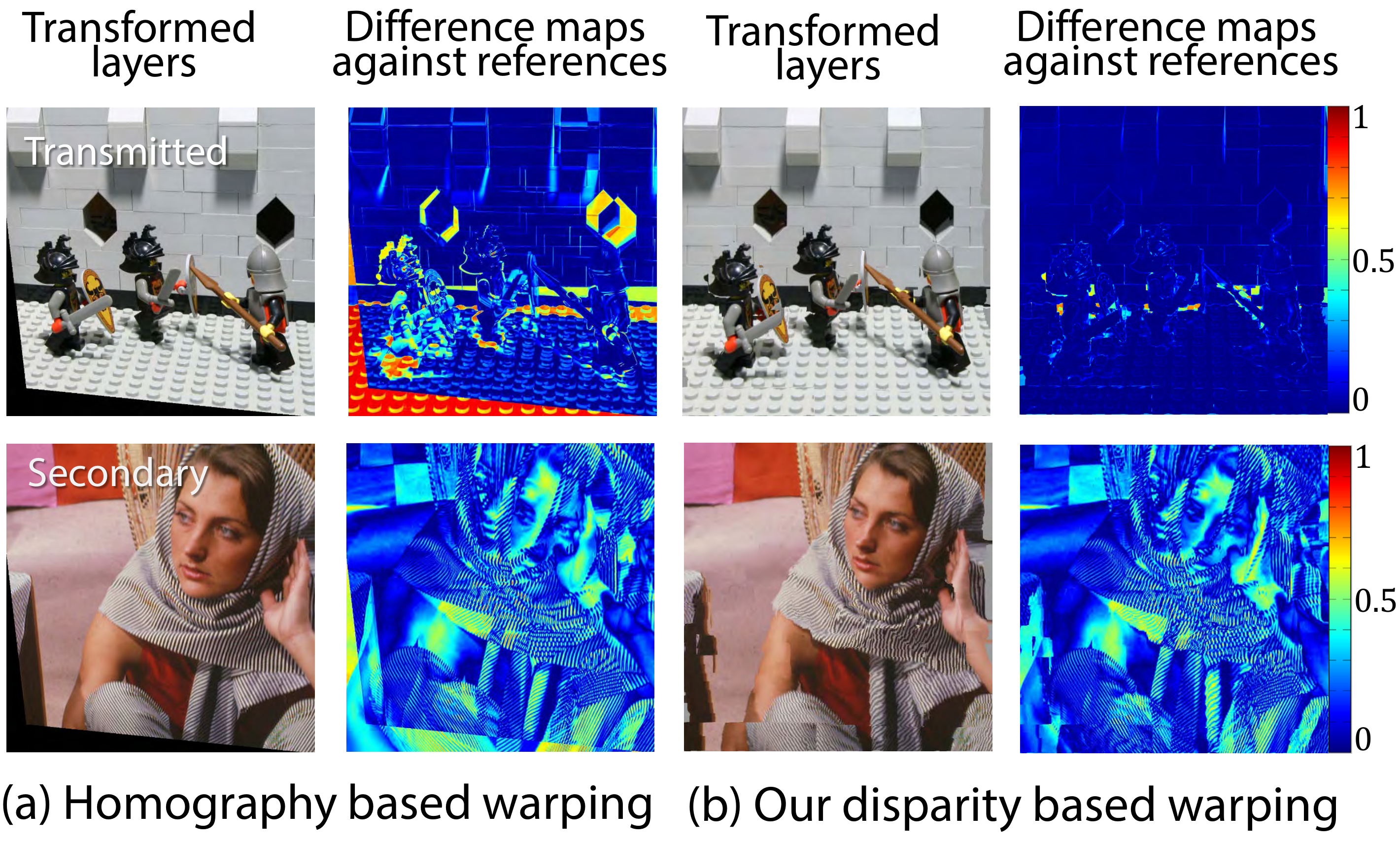}
   \end{center}
   \caption{Warping both transmitted and secondary layers using (a) homography vs. (b) disparity map. Disparity map produces more consistency than homography on the transmitted layer. Both transformations produce high incoherence on the secondary layer. }
   { \label{fig:depth}}
   \end{figure}


   \begin{figure*}
   \begin{center}
   \includegraphics[height=12cm]{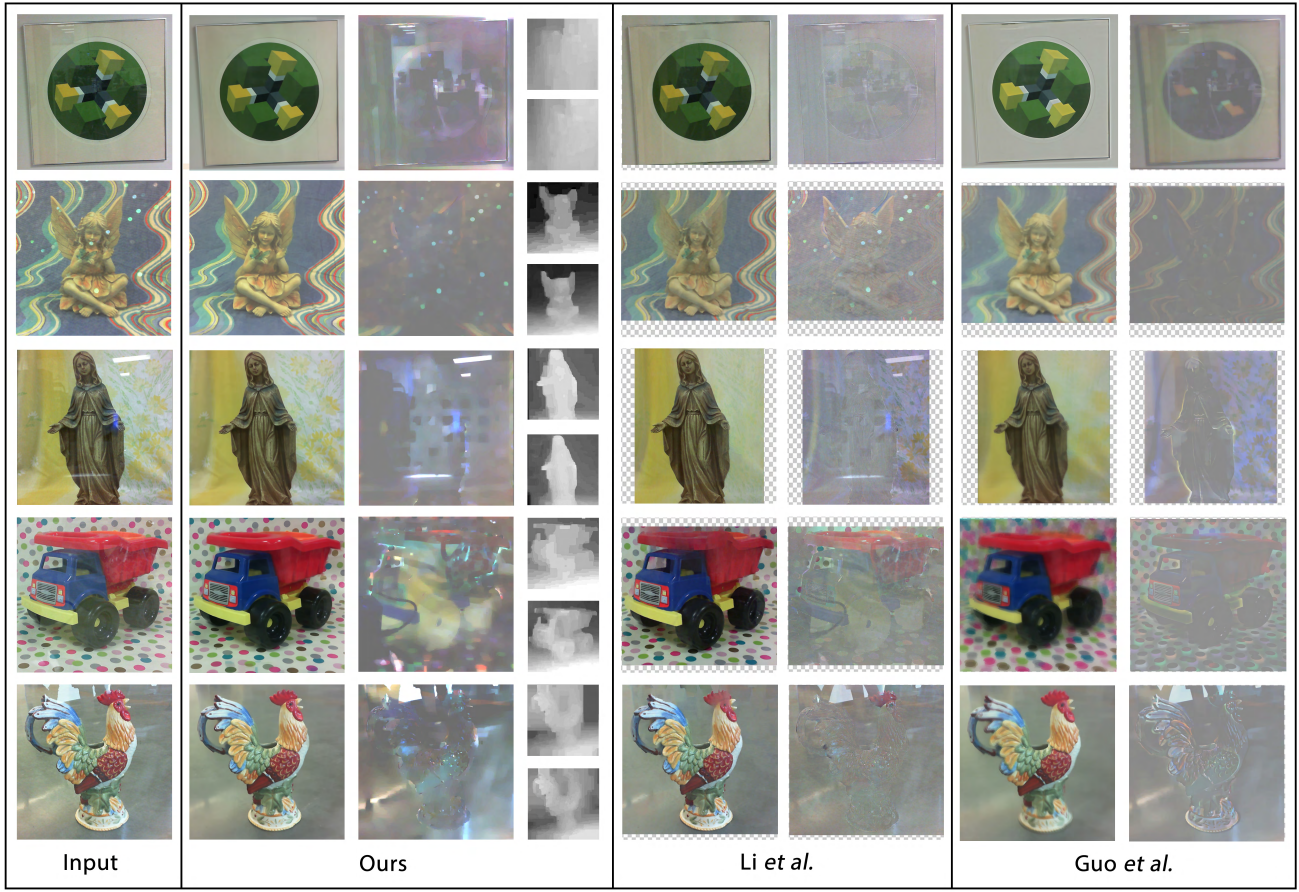}
   \end{center}
   \caption{Results on real scenes. From top to bottom: capturing a painting within a glass frame (row 1); a figurine behind a translucent layer of cloth (row 2); a copper statue, a plastic toy and a ceramic vase behind glass (last 3 rows). For each technique, we show recovered transmitted and secondary layers. Note that \protect\cite{li2013exploiting} (column 3) and \protect\cite{guo2014robust} (column 4) crop the original image and the reflection layer's contrast has been boosted for better visualization.
}
   { \label{fig:real}}
   \end{figure*}


   \begin{figure}
   \begin{center}
   \includegraphics[height=6 cm]{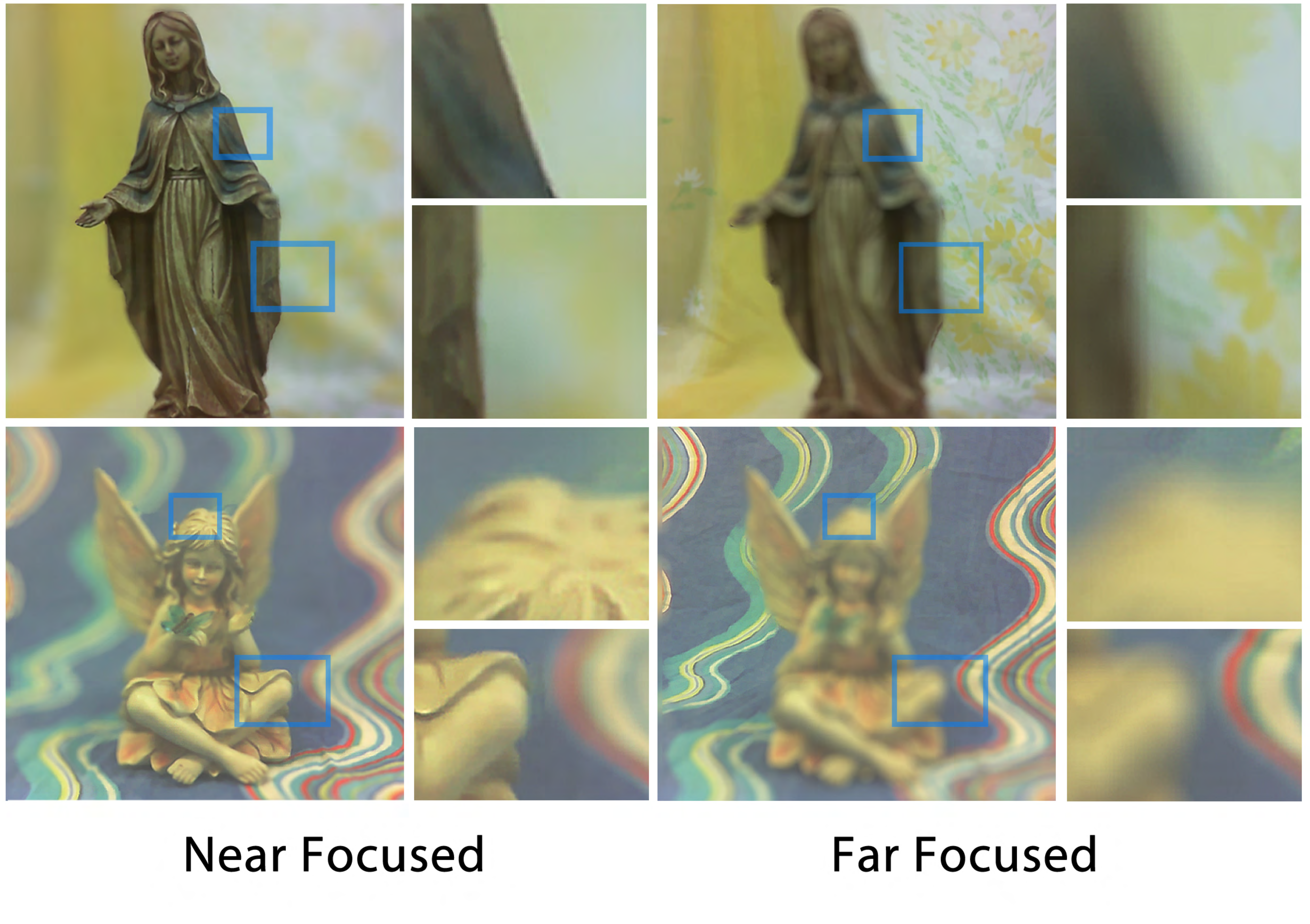}
   \end{center}
   \caption{Refocusing results. We demonstrate depth-guided refocusing using the depth map and transmitted layer image recovered by our algorithm. Close-ups show that color and depth boundaries are well-aligned.}
   { \label{fig:refocus}}
   \end{figure}


\section{Experiments}
We have conducted experiments on both synthetic and real data. All experiments are conducted on an Intel i7 PC (3.2GHz CPU, 16GB RAM) with the same set of parameters. We compared our results to two state-of-the-art techniques \cite{li2013exploiting} and \cite{guo2014robust}, by using the authors' source code with default parameters.

We first add synthetic reflections by superposing an additional layer to the Stanford LF images \cite{StanfordLightField}. The resolution of the synthetic images is of $1024 \times 1024$ and the motion of the additive layer is set to 20 pixels between adjacent views opposite to camera motion. Fig.~\ref{fig:synthetic} shows that our technique outperforms these alternative solutions in both accuracy and visual quality. This illustrates the importance of recovering the 3D shape of the transmitted layer. The multi-image technique of \cite{guo2014robust} uses homography (i.e., planes) as priors to register multiple images onto a common viewpoint. In our examples (e.g., the Stanford Bunny), the transmitted layer is non-planar and exhibits complex depth variations. As a result, \cite{guo2014robust} produces relatively large errors and ghosting artifacts due to image misalignment. In contrast, our technique has significantly less artifacts while recovering a relatively high quality disparity map. To illustrate the limitation of homography in transforming 3D scenes, we compare the transformed layers shown in Fig.~\ref{fig:depth}. Disparity based warping produces more consistency than homography on the transmitted layer.

The technique of \cite{li2013exploiting} is most similar to ours. It also models the transformation of the transmitted layers across different views as a flow field and uses SIFT flow for image warping. Therefore it is expected to better handle non-planar transmitted layer as shown in column 3 in Fig.~\ref{fig:synthetic}. However, it computes the flow field only once (at the beginning). Consequently, the separation quality is heavily reliant on the quality of flow initialization. For example, the bunny on the transmittance layer appears blurred in Fig.~\ref{fig:synthetic} since the initial disparity estimation is erroneous.

By comparison, our technique incorporates disparity estimation and layer separation into an iterative joint optimization framework. The benefits of our technique can be seen in Fig.~\ref{fig:synthetic}, with better detail recovery and better overall quality of layer separation.

For real experiments, we need to capture LF images with a reasonable baseline between adjacent viewpoints. We did not use the Lytro \cite{lytro} because it has an ultra-small baseline that limits its working range to only about 6 inches, whereas existing camera arrays are too bulky for practical use. We built our own portable LF array consisting of $9$ Microsoft LifeCam HD-6000 USB cameras on a 3D-printed grid (Fig.~\ref{fig:sample_lf}). The resolution of each camera is $2560 \times 1440$, and the baseline can be set to either 1, 2, or 3 inches. To capture static scenes, we connect all cameras to a Keynice H1088 10-port hub powered by an Anker Astro Pro2 external battery pack. A single HP Stream 7 tablet is used to trigger individual cameras and store data. It takes around 1 second to take all 9 shots at full resolution. To capture dynamic scenes, we connect the cameras to a workstation equipped with 3 PCI-E USB 3.0 adaptors, each having 4 dedicated 5Gbps channels. This configuration allows us to record HD (720p) LF videos at 30 fps. We pre-calibrate the camera using the technique described in \cite{zhang2000flexible}.

For validation, we captured some scenes with a reflective layer and others with a translucent layer. We first capture a LF of a painting within a glass frame using the 3-inch baseline. This is a typical problem that \cite{guo2014robust} aims to solve. Our method produces comparable results. However, it is worth noting that \cite{guo2014robust} requires users to manually find four corresponding corners in a view for computing the homography. We instead automatically compute the disparity map without any user input. In the second example, we capture a figurine behind a translucent layer of cloth using the 1-inch baseline. Our method is able to reliably recover the 3D geometry of the figurine as well as remove the effect of cloth layer. To use \cite{guo2014robust}, we select four feature points on the images and approximate a homography for warping the images. Their results exhibit clear visual artifacts due to their inability to account for arbitrary depth variation.

Next, we capture three objects made of different materials behind a reflective glass. This emulates the museum setting of photographing 3D artifacts. These objects, especially the toy truck, have clear depth variations and the parallax across the LF views violates the homography model. Consequently, both the recovered transmitted layer and the secondary layer from \cite{guo2014robust} exhibit ghosting artifacts due to misalignment of views. The technique of \cite{li2013exploiting} partially reduces these artifacts as initial SIFT flow better register the images. However, the SIFT flow still has large deviation from the actual disparity map and their results exhibit artifacts on heavily saturated regions due to misalignment.

\begin{figure}
   \begin{center}
   \includegraphics[height=6 cm]{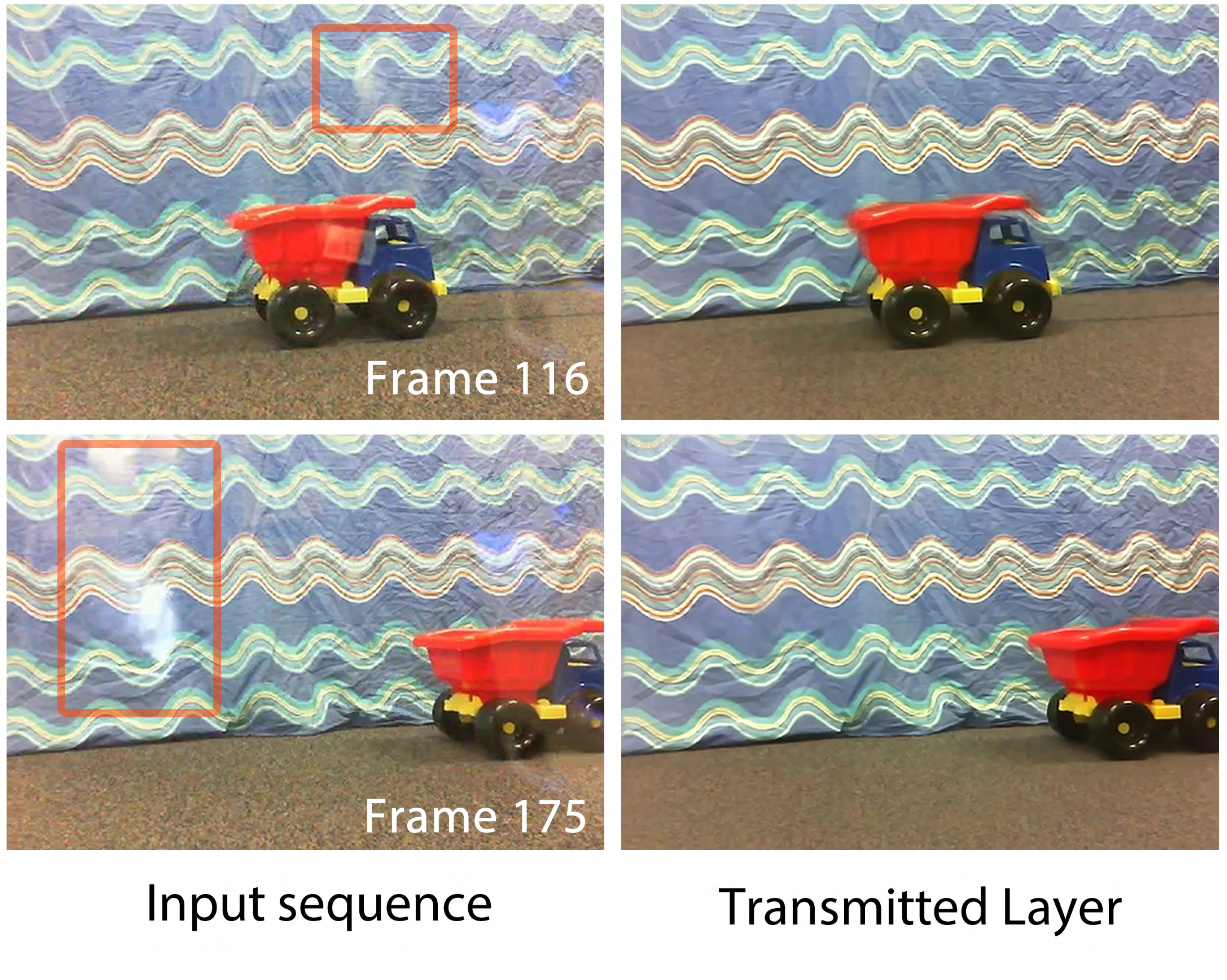}
   \end{center}
   \caption{Results on a dynamic scene with moving objects in both layers. \emph{Left:} input image frames, changes in reflection are highlighted. \emph{Right:} transmitted layer recovered by the proposed method.}
   { \label{fig:video}}
   \end{figure}

Our technique is able to generate better results. More importantly, with the help of the disparity map, we are able to align the views and eliminate most of the reflection layers while preserving fine geometric details and texture, as seen in Fig.~\ref{fig:real}. Our layer separation solution also produces a high quality 3D depth map, with which we can perform IBR effects such as depth-guided refocusing (Fig.~\ref{fig:refocus}) on the transmitted layer. Fig.~\ref{fig:video} shows our results on a dynamic scene with a toy truck moving behind glass. The bottom row shows results of removing the fast moving reflection. To the best of our knowledge, our solution is the first to perform reliable layer separation on dynamic scenes.

We examined our LF camera in a variety of environments, and found that the 1-inch baseline provides enough view changes for almost all practical scenes that are 4-6 feet away. Also, a $3\times3$ LF is sufficient for nearly all cases. More views will further improve the low-rank constraint in RPCA optimization but is also more computationally expensive. Our method takes about 7 minutes on average to process one LF video frame (containing 9 views at a resolution of $640 \times 480$). The code of \cite{guo2014robust} takes about 3 minutes to finish a image sequence of the same size. The author of \cite{li2013exploiting} reports a running time of about 5 minutes for a $500 \times 400$ image sequence containing up to 5 images.

As with previous techniques, we assume that the transmitted layer is dominant with the contribution of the secondary layer being relatively small. This ensures that the SIFT flow algorithm will mostly choose feature points from the transmitted layer to produce mostly correct warping. If the assumption is violated, the detected feature points will come from a mixed pool of two layers. Since our iterative refinement process is local, it may not be able to overcome the large errors.

\begin{figure}
   \begin{center}
   \includegraphics[height=5.6 cm]{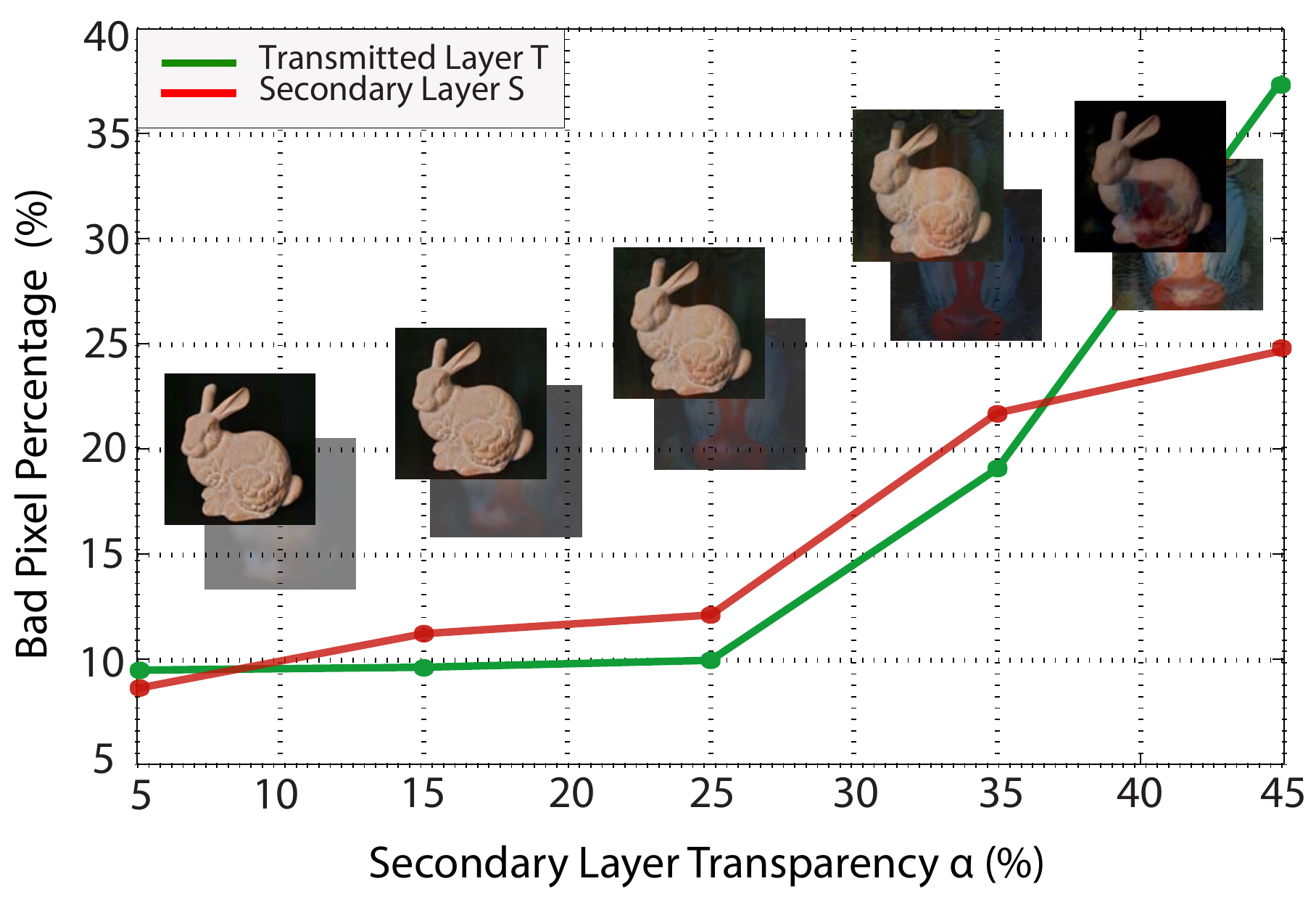}
   \end{center}
   \caption{Reconstruction accuracy vs. Transparency. We apply different blending parameters for combining the transmitted and secondary layers on the Stanford Bunny scene. We use a light field $3\times3$ views with a resolution of $1024 \times 1024$ and compute the percentage of incorrectly recovered pixels for both layers.}
   { \label{fig:alpha}}
   \end{figure}

We experimented on a synthetic scene dataset where we control the blending of the two layers with a blending parameter $\alpha$. We apply our layer separation technique for different values of $\alpha$. We compute the percentage of incorrectly recovered pixels in both layers where we use 0.1 (for intensity range [0, 1]) as the threshold to determine if a recovered pixel is incorrect. Fig.~\ref{fig:alpha} shows the layer separation accuracy versus $\alpha$. For small $\alpha$ (e.g., in range $[0,25]\%$), we are able to obtain good results. The performance significantly degrades when $\alpha$ is above $35\%$ and our algorithm fails when $\alpha$ is above $50\%$.

\section{Conclusion}
We have presented a novel technique that automatically separate the transmitted and secondary layers. At the core of our technique is the use of light field imaging to acquire multi-view images. With approximate scene depth of the transmitted layer, we can warp all light field views to the reference view to form an image stack. The corresponding transmitted stack is expected to be of low rank, while the secondary layer is of low coherence and hence sparse. We start with SIFT flow to generate the initial depth map and then apply an iterative optimization scheme based on Robust PCA (RPCA) for layer separation and depth map refinement. It is worth noting that our technique handles dynamic scenes (e.g., removing reflections from video), which would be almost impossible for traditional methods using an unstructured collection of viewpoints.

An implicit assumption of our technique is that the transmitted layer is predominant so that SIFT flow can produce a reliable initial estimation of the disparity map of the transmitted layer. We plan to investigate the structure of the secondary layer to relax this assumption. We would also like to try our technique on the Pelican \cite{Pelicanimaging} or Light \cite{Lightstartup} mobile LF camera which is expected to be on the market soon and compare their results with those from our light field setup. Another interesting direction is to estimate 3D shape of the secondary layer as well, by reformulating our problem using two unknown disparity maps.


{\small
\bibliographystyle{ieee}
\bibliography{egbib}

\begin{thebibliography}{10}\itemsep=-1pt

\bibitem{Pelicanimaging}
Pelican imaging mobile camera array.
\newblock \url{https://www.pelicanimaging.com}, 2013 (Accessed on March, 2015).

\bibitem{Lightstartup}
Light mobile camera array.
\newblock \url{https://light.co/}, 2015 (Accessed on March, 2015).

\bibitem{adelson1992single}
E.~H. Adelson and J.~Y.~A. Wang.
\newblock Single lens stereo with a plenoptic camera.
\newblock {\em IEEE TPAMI}, 14(2):99--106, 1992.

\bibitem{agrawal2005removing}
A.~Agrawal, R.~Raskar, S.~K. Nayar, and Y.~Li.
\newblock Removing photography artifacts using gradient projection and
  flash-exposure sampling.
\newblock In {\em ACM TOG}, volume~24, pages 828--835, 2005.

\bibitem{candes2011robust}
E.~J. Cand{\`e}s, X.~Li, Y.~Ma, and J.~Wright.
\newblock Robust principal component analysis?
\newblock {\em Journal of the ACM}, 58(3):11, 2011.

\bibitem{ChenCVPR2014}
C.~Chen, H.~Lin, S.~B.~K. Z.~Yu, and J.~Yu.
\newblock Light field stereo matching using bilateral statistics of surface
  cameras.
\newblock In {\em CVPR}, 2014.

\bibitem{StanfordLightField}
S.~U. Computer Graphics~Laboratory.
\newblock The (new) stanford light field archive.
\newblock \url{http://lightfield.stanford.edu/}, 2008 (Accessed on March,
  2015).

\bibitem{farid1999separating}
H.~Farid and E.~H. Adelson.
\newblock Separating reflections from images by use of independent component
  analysis.
\newblock {\em JOSA A}, 16(9):2136--2145, 1999.

\bibitem{gai2012blind}
K.~Gai, Z.~Shi, and C.~Zhang.
\newblock Blind separation of superimposed moving images using image
  statistics.
\newblock {\em IEEE TPAMI}, 34(1):19--32, 2012.

\bibitem{guo2014robust}
X.~Guo, X.~Cao, and Y.~Ma.
\newblock Robust separation of reflection from multiple images.
\newblock In {\em CVPR}, pages 2195--2202, 2014.

\bibitem{heber2014shape}
S.~Heber and T.~Pock.
\newblock Shape from light field meets robust pca.
\newblock In {\em ECCV}, pages 751--767. 2014.

\bibitem{HeberICEMM2013}
S.~Heber, R.~Ranftl, and T.~Pock.
\newblock Variational shape from light field.
\newblock In {\em International Conference on Energy Minimization Methods in
  Computer Vision and Pattern Recognition}, 2013.

\bibitem{hirschmuller2008stereo}
H.~Hirschmuller.
\newblock Stereo processing by semiglobal matching and mutual information.
\newblock {\em IEEE TPAMI}, 30(2):328--341, 2008.

\bibitem{kim2013scene}
C.~Kim, H.~Zimmer, Y.~Pritch, A.~Sorkine-Hornung, and M.~H. Gross.
\newblock Scene reconstruction from high spatio-angular resolution light
  fields.
\newblock {\em ACM Trans. Graph.}, 32(4):73, 2013.

\bibitem{kong2011high}
N.~Kong, Y.-W. Tai, and S.~Y. Shin.
\newblock High-quality reflection separation using polarized images.
\newblock {\em IEEE TIP}, 20(12):3393--3405, 2011.

\bibitem{levin2007user}
A.~Levin and Y.~Weiss.
\newblock User assisted separation of reflections from a single image using a
  sparsity prior.
\newblock {\em IEEE TPAMI}, 29(9):1647--1654, 2007.

\bibitem{levin2004separating}
A.~Levin, A.~Zomet, and Y.~Weiss.
\newblock Separating reflections from a single image using local features.
\newblock In {\em CVPR}, volume~1, pages I--306, 2004.

\bibitem{li2013exploiting}
Y.~Li and M.~S. Brown.
\newblock Exploiting reflection change for automatic reflection removal.
\newblock In {\em ICCV}, pages 2432--2439, 2013.

\bibitem{Lippermann1908}
G.~Lippmann.
\newblock La photographie intgrale.
\newblock {\em Comptes-Rendus,Acadmie des Sciences}, 146:446--451, 1908.

\bibitem{liu2011sift}
C.~Liu, J.~Yuen, and A.~Torralba.
\newblock Sift flow: Dense correspondence across scenes and its applications.
\newblock {\em IEEE TPAMI}, 33(5):978--994, 2011.

\bibitem{lumsdaine2009focused}
A.~Lumsdaine and T.~Georgiev.
\newblock The focused plenoptic camera.
\newblock In {\em IEEE ICCP}, pages 1--8, 2009.

\bibitem{lytro}
Lytro.
\newblock Lytro camera.
\newblock \url{https://www.lytro.com}, 2012 (Accessed on March, 2015).

\bibitem{maeno2013light}
K.~Maeno, H.~Nagahara, A.~Shimada, and R.-i. Taniguchi.
\newblock Light field distortion feature for transparent object recognition.
\newblock In {\em CVPR}, pages 2786--2793, 2013.

\bibitem{ng2005light}
R.~Ng, M.~Levoy, M.~Br{\'e}dif, G.~Duval, M.~Horowitz, and P.~Hanrahan.
\newblock Light field photography with a hand-held plenoptic camera.
\newblock {\em Computer Science Technical Report CSTR}, 2(11), 2005.

\bibitem{scharstein2002taxonomy}
D.~Scharstein and R.~Szeliski.
\newblock A taxonomy and evaluation of dense two-frame stereo correspondence
  algorithms.
\newblock {\em IJCV}, 47(1-3):7--42, 2002.

\bibitem{schechner2000separation}
Y.~Y. Schechner, N.~Kiryati, and R.~Basri.
\newblock Separation of transparent layers using focus.
\newblock {\em IJCV}, 39(1):25--39, 2000.

\bibitem{schechner2000polarization}
Y.~Y. Schechner, J.~Shamir, and N.~Kiryati.
\newblock Polarization and statistical analysis of scenes containing a
  semireflector.
\newblock {\em JOSA A}, 17(2):276--284, 2000.

\bibitem{sinha2012image}
S.~N. Sinha, J.~Kopf, M.~Goesele, D.~Scharstein, and R.~Szeliski.
\newblock Image-based rendering for scenes with reflections.
\newblock {\em ACM Trans. Graph.}, 31(4):100, 2012.

\bibitem{szeliski2000layer}
R.~Szeliski, S.~Avidan, and P.~Anandan.
\newblock Layer extraction from multiple images containing reflections and
  transparency.
\newblock In {\em CVPR}, volume~1, pages 246--253, 2000.

\bibitem{taodepth2014}
M.~W. Tao, T.-C. Wang, J.~Malik, and R.~Ramamoorthi.
\newblock Depth estimation for glossy surfaces with light-field cameras.
\newblock In {\em ECCV Workshop on Light Fields for Computer Vision}. 2014.

\bibitem{tsin2006stereo}
Y.~Tsin, S.~B. Kang, and R.~Szeliski.
\newblock Stereo matching with linear superposition of layers.
\newblock {\em IEEE TPAMI}, 28(2):290--301, 2006.

\bibitem{wanner2013variational}
S.~Wanner and B.~Goldluecke.
\newblock Variational light field analysis for disparity estimation and
  super-resolution.
\newblock 2013.

\bibitem{YuICCV2013}
Z.~Yu, X.~Guo, H.~Ling, A.~Lumsdaine, and J.~Yu.
\newblock Line-assisted light field triangulation and stereo matching.
\newblock In {\em ICCV}, 2013.

\bibitem{zhang2000flexible}
Z.~Zhang.
\newblock A flexible new technique for camera calibration.
\newblock {\em Pattern Analysis and Machine Intelligence, IEEE Transactions
  on}, 22(11):1330--1334, 2000.

\end{thebibliography}
}

\end{document}